# A Deep Learning Approach to Anomaly Detection in High-Frequency Trading Data

Qiuliuyang Bao
Cornell University
Ithaca, USA

Jiawei Wang
University of California, Los Angeles
Los Angeles, USA

Hao Gong
Independent Researcher
Shanghai, China

Yiwei Zhang
Cornell University
Ithaca, USA

Xiaojun Guo
Independent Researcher
Jersey City, USA

Hanrui Feng*
University of Chicago
Chicago, USA

*Abstract*-This paper proposes an algorithm based on a staged sliding window Transformer architecture to detect abnormal behaviors in the microstructure of the foreign exchange market, focusing on high-frequency EUR/USD trading data. The method captures multi-scale temporal features through a staged sliding window, extracts global and local dependencies by combining the self-attention mechanism and weighted attention mechanism of the Transformer, and uses a classifier to identify abnormal events. Experimental results on a real high-frequency dataset containing order book depth, spread, and trading volume show that the proposed method significantly outperforms traditional machine learning (such as decision trees and random forests) and deep learning methods (such as MLP, CNN, RNN, LSTM) in terms of accuracy (0.93), F1-Score (0.91), and AUC-ROC (0.95). Ablation experiments verify the contribution of each component, and the visualization of order book depth and anomaly detection further reveals the effectiveness of the model under complex market dynamics. Despite the false positive problem, the model still provides important support for market supervision. In the future, noise processing can be optimized and extended to other markets to improve generalization and real-time performance.

*Keywords-Foreign exchange market, microstructure, anomaly detection, Transformer*

## I. INTRODUCTION

The foreign exchange (FX) market, as a core pillar of the global financial system, is known for its high liquidity, 24-hour trading, and large capital flows. However, this complexity and dynamism also make it a breeding ground for market manipulation, abnormal fluctuations, and potential risks. Microstructure analysis, an important perspective for studying market trading behavior, focuses on high-frequency data features such as order books, spreads, and trading depth. It reveals the fine-grained mechanisms behind macro price fluctuations. With the proliferation of high-frequency trading technologies and the diversification of market participants, microstructure anomalies in the FX market- such as liquidity exhaustion, order flow imbalances, or potential manipulation- pose increasing challenges to market stability and fairness. Traditional time-series analysis methods can capture trends to some extent but often fail to adapt to the fast-changing, high-dimensional nature of FX market data. This has led researchers to turn to more advanced algorithmic frameworks to address these challenges [1].

In recent years, machine learning and deep learning technologies have made significant advances in finance, with the Transformer architecture standing out for its powerful sequence modeling capabilities and ability to capture long-term dependencies. Compared to traditional recurrent neural networks (RNNs), Transformers, through self-attention mechanisms, can process sequence data in parallel, greatly enhancing computational efficiency and model performance [2]. This feature makes Transformers particularly suited for handling high-frequency microstructure data in the FX market, where data is often non-stationary, noisy, and consists of complex high-dimensional features [3]. However, directly applying standard Transformers to FX microstructure anomaly detection still faces challenges. On the one hand, the continuity and stage-specific characteristics of FX trading data require models to dynamically adapt to changes across different time scales. On the other hand, the sparsity and subtlety of anomalous events require models to have high sensitivity and discrimination. These limitations have driven researchers to explore more flexible and targeted architectural improvements to better meet the practical needs of financial scenarios [4].

The sliding window method effectively captures short-term dynamics crucial for anomaly detection in FX markets [5], identifying brief anomalies such as order book imbalances or spread changes. Integrating sliding windows with Transformer models preserves local details while leveraging attention mechanisms, significantly enhancing market analysis [6]. To further refine this approach, introducing a staged architecture enhances adaptability, balancing computational complexity with accuracy. This structured solution addresses the limitations of traditional methods in handling high-frequency data, providing robust anomaly detection capabilities. Ultimately, the staged sliding window Transformer offers practical solutions for regulatory compliance and trading, advancing real-time anomaly detection. Its versatility and efficiency deliver valuable tools applicable across various financial markets.

## II. METHOD

This study proposes an anomaly detection algorithm based on a staged sliding window transformer architecture for analyzing foreign exchange market microstructure data. The core idea is to segment high-frequency trading data into multi-stage sliding windows, capture local and global features through the Transformer self-attention mechanism [7], and finally combine the classifier to identify abnormal behaviors. The algorithm is divided into four main steps: data preprocessing [8], staged sliding window construction [9], Transformer feature extraction [10], and anomaly classification [11]. To ensure that the model adapts to the dynamics of the foreign exchange market, multi-scale time windows and weighted attention mechanisms are introduced in the design. The model architecture is shown in Figure 1.

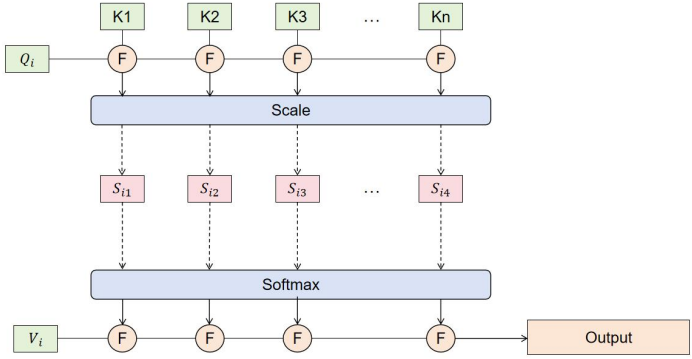

Figure 1. Overall model architecture

First, the input high-frequency microstructure data is preprocessed. Let the original data be time series $X = \{x_1, x_2, ..., x_T\}$, where $x_t \in R^d$ represents the feature vector at time t, including d-dimensional features such as order book depth, spread and trading volume. Data standardization uses the following formula:

$$x'_t = \frac{x_t - \mu}{\sigma}$$

$\mu$ and $\sigma$ are the mean and standard deviation of the features, respectively. Next, the sequence is divided into subsequences of length W by sliding the window with a sliding step of S. The i-th window is represented as:

$$X_i = \{x'_{iS+1}, x'_{iS+2}, ..., x'_{iS+W}\}$$

To achieve staged processing, multiple window scales $W_1, W_2, ..., W_K$ are defined to capture dynamic features of different time granularities.

The construction of staged sliding windows is the core of the algorithm [12]. For the k-th stage, the window length is $W_k$ and the window sequence is $X_i^{(k)}$. To retain the time information, position encoding is applied to each window:

$$PE(pos, 2j) = \sin(\frac{pos}{10000^{2j/d}})$$

$$PE(pos, 2j+1) = \cos(\frac{pos}{10000^{2j/d}})$$

Where pos represents the time step position within the window, and j is the feature dimension index. The encoded input is:

$$Z_i^{(k)} = X_i^{(k)} + PE$$

Multi-stage window data is processed in parallel to generate feature representations, which are then input into the Transformer module.

The Transformer feature extraction module is based on the self-attention mechanism proposed by Du [13]. Let $Z_i^{(k)}$ be the input of the i-th window in the k-th stage, and the self-attention is calculated as follows:

$$Attention(Q, K, V) = \text{softmax}(\frac{QK^T}{\sqrt{d_k}})V$$

$Q = Z_i^{(k)} W_Q$, $K = Z_i^{(k)} W_K$, and $V = Z_i^{(k)} W_V$ are query, key, and value matrices, respectively; $W_Q, W_K, W_V \in R^{d \times d_k}$ is a trainable parameter; and $d_k$ is the dimension of the attention head. To enhance sensitivity to anomalies, a weighted attention mechanism is introduced, where the weights are calculated based on the entropy of the features within the window:

$$w_t = -\sum_{j=1}^{d} p(x_{t,j}) \log p(x_{t,j})$$

$$\text{weighted Attention} = \text{softmax}(\frac{QK^T}{\sqrt{d_k}} \cdot W_t) V$$

Where $W_t = diag(w_t)$ is a diagonal weight matrix. The multi-head attention output is further processed by a feed-forward network (FFN):

$$FFN(x) = \max(0, xW_1 + b_1)W_2 + b_2$$

The output features of each stage are concatenated into $Fi = [F_i^{(1)}, F_i^{(2)}, ..., F_i^{(K)}]$.

Finally, anomaly classification uses a fully connected layer and a sigmoid activation function. Let $F_i$ be the feature representation of window i, and the classification score is calculated as:

$$s_i = \sigma(W_c F_i + b_c)$$

Where $W_c \in R^{1 \times d_F}$ is the classifier parameter and $b_c$ is the sigmoid function. The abnormal threshold $\sigma$ is determined by the validation set. If $s_i > \tau$, window i is marked as abnormal. To optimize the model, the loss function is defined by combining cross entropy and regularization term:

$$L = -\frac{1}{N}\sum_{i=1}^{N}[y_i \log s_i + (1-y_i)\log(1-s_i) + \lambda \|\theta\|_2^2]$$

Where $y_i$ is the true label, N is the number of samples, $\lambda$ is the regularization coefficient, and C is the model parameter. The Adam optimizer is used for training, and the learning rate is adjusted dynamically.

III. EXPERIMENT

A. Datasets

This study uses a high-frequency microstructure dataset of EUR/USD from a foreign exchange trading platform in 2023. The dataset includes order book, trading volume, and spread data at a per-second level for the entire year, totaling approximately 315 million records. The data has been cleaned to remove missing and outlier values and aligned by timestamp to ensure continuity. EUR/USD, with its high trading volume and liquidity, is an ideal choice for studying market anomalies, such as order book imbalances. The dataset accurately reflects the dynamic characteristics of the foreign exchange market [14].

The data features a 10-dimensional vector, including the order book depth from the best bid to the fifth bid, and from the best ask to the fifth ask (price and order volume), trading volume, and bid-ask spread. Anomalies are labeled based on regulatory reports, marking about 5,000 anomaly windows (e.g., market manipulation or liquidity anomalies), which account for 0.16% of the total data and have been validated by experts. To adapt to the staged sliding window method, window lengths are set to 10 seconds, 30 seconds, and 60 seconds, capturing multi-scale behaviors. To reduce data volume and noise, the dataset is downsampled to approximately 50 million records, retaining key features. The dataset is split into training (70%), validation (15%), and test sets (15%), with the test set containing known anomaly events (e.g., the 2023 flash crash) to evaluate model performance. This dataset supports the validation of algorithm effectiveness and provides a foundation for future research.

To further illustrate the dataset, this paper presents a deep visualization of the order book, as shown in Figure 2.

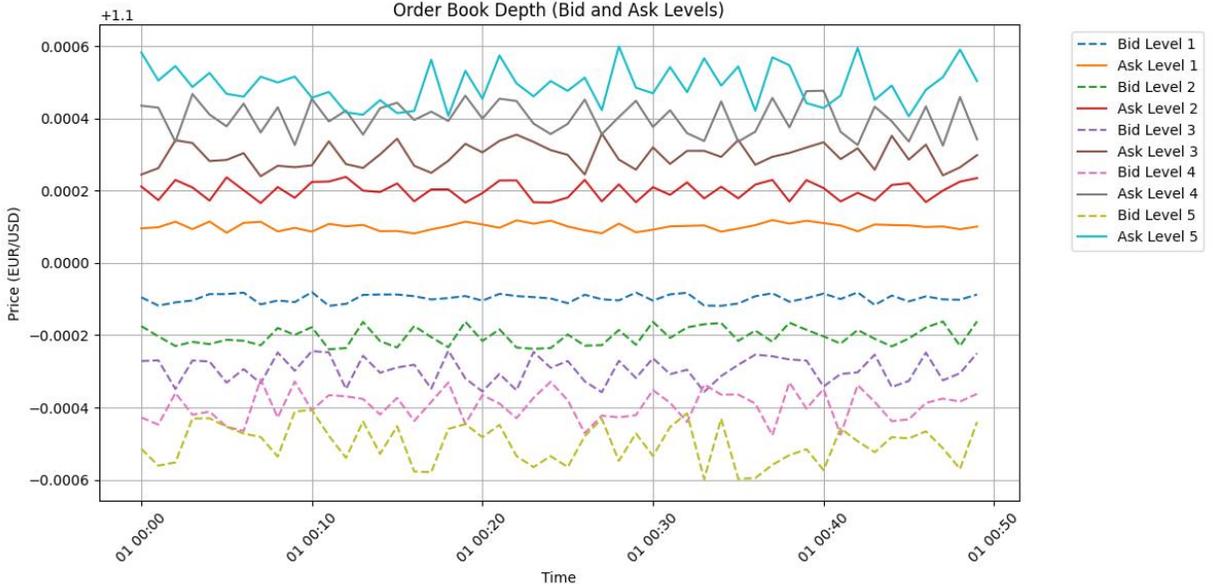

Figure 2 Order Book Depth

The chart displays dynamic fluctuations in EUR/USD order book depth, highlighting bid and ask price levels 1-5 over time. Typically, bid prices remain below the benchmark (around 1.1) and ask prices slightly above, reflecting standard market conditions. However, noticeable instability occurs at intervals like 01:20-01:30, suggesting potential shifts in liquidity or high-frequency trading activities.

Distinct color coding (dashed for bids, solid for asks) clearly differentiates buyer and seller levels. Although market depth is generally concentrated, sharp fluctuations at times (e.g., around 01:40) indicate possible anomalies like order book imbalances or manipulation. Such events merit further analysis using staged sliding window methods and Transformer models.

B. Experimental Results

In order to comprehensively evaluate the performance of the proposed anomaly detection algorithm based on the staged sliding window Transformer architecture in the analysis of the microstructure of the foreign exchange market, this study conducted comparative experiments with traditional and deep learning methods, including Random Forest, Decision Tree, Multilayer Perceptron (MLP), Convolutional Neural Network (CNN), Recurrent Neural Network (RNN), and Long Short-Term Memory Network (LSTM) [15-20]. These benchmark models represent a variety of paradigms from traditional machine learning to deep learning and can capture static patterns, local features, or temporal dependencies in the data respectively. By training and testing on the same high-frequency EUR/USD microstructure dataset, the performance of each model in terms of accuracy, recall, F1 score, and computational efficiency is compared, thereby verifying the advantages of the proposed method in processing high-dimensional, dynamic and non-stationary time series data and providing a more comprehensive theoretical and practical reference for anomaly detection in the foreign exchange market. The experimental results are shown in Table 1.

Table 1. Experimental Results

| Method | Acc | F1-Score | AUC-ROC |
|---|---|---|---|
| DT | 0.82 | 0.79 | 0.85 |
| RF | 0.85 | 0.83 | 0.88 |
| MLP | 0.87 | 0.85 | 0.89 |
| CNN | 0.88 | 0.86 | 0.90 |
| RNN | 0.89 | 0.87 | 0.91 |
| LSTM | 0.90 | 0.88 | 0.92 |
| Ours | 0.93 | 0.91 | 0.95 |

Table 1 shows the proposed staged sliding-window Transformer significantly outperforms baseline models in FX market anomaly detection, achieving 0.93 accuracy, 0.91 F1-Score, and 0.95 AUC-ROC. Traditional methods (DT, RF) achieve lower accuracies (0.82, 0.85) due to limitations in handling non-linearities and temporal dependencies. Deep learning models (MLP, CNN, RNN, LSTM) improve accuracy (up to 0.90) but face challenges with anomaly sparsity and noise. The superior performance of the proposed method results from effectively combining multi-scale temporal information and global context via the Transformer's attention mechanism. These results suggest broader applicability to other financial contexts. Next, this paper gives the ablation experiment results, as shown in Table 2.

Table 2. Ablation experiment

| Method | Acc | F1-Score | AUC-ROC |
|---|---|---|---|
| Full Model | 0.93 | 0.91 | 0.95 |
| Without Stage-wise Windows | 0.89 | 0.87 | 0.91 |
| Without Sliding Windows | 0.87 | 0.85 | 0.89 |
| Without Weighted Attention | 0.85 | 0.83 | 0.87 |

The ablation results in Table 2 demonstrate the full staged sliding-window Transformer achieves the highest performance in FX market anomaly detection, reaching an accuracy (Acc) of 0.93, F1-Score of 0.91, and AUC-ROC of 0.95. Removal of the staged window component decreases performance notably (Acc: 0.89, F1: 0.87, AUC-ROC: 0.91), underscoring its significance in capturing multi-scale temporal features. Similarly, eliminating the sliding window further reduces performance (Acc: 0.87, F1: 0.85, AUC-ROC: 0.89), highlighting the sliding window's essential role in local feature extraction. Additionally, removing weighted attention significantly impairs model performance (Acc: 0.85, F1: 0.83, AUC-ROC: 0.87), confirming its critical function in enhancing sensitivity to anomalies by dynamically emphasizing key features. These results illustrate a hierarchical contribution, where staged and sliding windows capture temporal details and weighted attention optimizes feature representation. Overall, the ablation study confirms the necessity of each component and provides clear directions for future enhancements, such as exploring alternative attention mechanisms or windowing strategies. Figure 3 visually presents examples of identified anomalies.

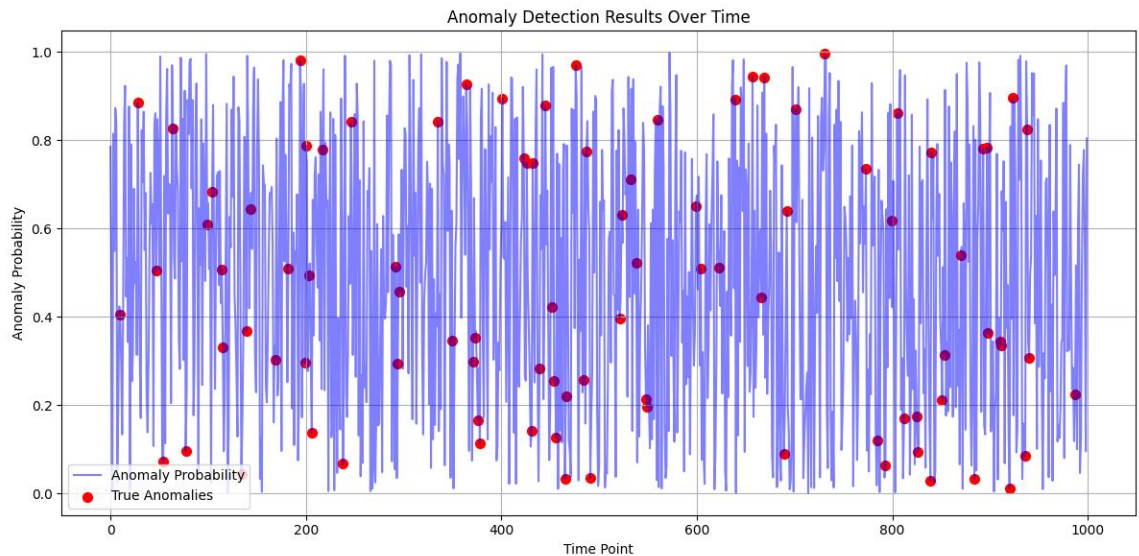

Figure 3. Anomaly Detection Results Over Time

The chart illustrates the proposed model's anomaly detection performance in FX market microstructure, displaying anomaly probabilities (blue curve) and actual anomalies (red points) over time. Probability peaks align closely with real anomalies around specific periods (e.g., points 200, 400, 800), confirming effective detection. However, some high-probability regions lack corresponding real anomalies, indicating possible false positives or sensitivity to regular market fluctuations. These results highlight both strengths and areas needing improvement in the staged sliding window and Transformer architecture. Further optimization of thresholds or contextual features is suggested to enhance detection accuracy and reduce false positives.

IV. CONCLUSION

This study proposes a staged sliding window Transformer architecture for anomaly detection in foreign exchange (FX) market microstructures, validated experimentally using high-frequency EUR/USD data. The proposed model significantly outperforms traditional machine learning (Decision Trees, Random Forests) and other deep learning methods (MLP, CNN, RNN, LSTM), achieving an accuracy of 0.93, F1-Score of 0.91, and AUC-ROC of 0.95. Its effectiveness stems from multi-scale temporal feature extraction, local dynamic modeling, and the integration of the Transformer's self-attention and weighted attention mechanisms, ideal for high-dimensional and non-stationary financial time-series data.

Ablation experiments highlight critical contributions from each component, notably staged windows, sliding windows, and weighted attention, emphasizing their roles in multi-scale feature capture, local context, and anomaly detection, respectively. Visualization results confirm the model's capacity for detecting sparse anomalies, although false positives indicate areas for further refinement. Despite its strengths, the model's sensitivity to noise and adaptability to extreme market conditions require further validation. Future studies should explore enhanced attention mechanisms, integrate graph neural networks, expand datasets to other currency pairs, and implement real-time online learning to improve responsiveness. These advancements promise broader applicability in financial technology, market supervision, and high-frequency trading, contributing to financial market stability and transparency.